\documentclass[letterpaper,conference, 10pt]{style/ieeeconf}

\IEEEoverridecommandlockouts    

\usepackage{amsthm}
\usepackage[T1]{fontenc}
\usepackage[latin9]{inputenc}
\usepackage{calc}
\usepackage{amssymb, amsmath}
\usepackage{graphicx}
\usepackage{xcolor}
\usepackage{mathtools}
\usepackage{cite}
\usepackage{hyperref}
\usepackage[capitalize]{cleveref}
\usepackage{subfigure}
\usepackage[noend]{algpseudocode}
\usepackage{algorithm}
\usepackage{multicol}
\usepackage{multirow}
\usepackage{titlesec}
\usepackage[font=footnotesize, skip=1ex, belowskip=-10pt]{caption}
\usepackage{flushend} 

\usepackage{style/perl_math}

\titlespacing{\subsection}{0pt}{5pt plus 0pt minus 2pt}{3pt plus 0pt minus 2pt}
\titlespacing{\subsubsection}{0pt}{5pt plus 0pt minus 2pt}{3pt plus 0pt minus 2pt}

\newtheoremstyle{nospace}{2pt}{1.5pt}{\itshape}{}{\bfseries}{:}{.5em}{}
\theoremstyle{nospace}
\newtheorem{assumption}{Assumption}
\newtheorem{definition}{Definition}

\begin{document}
\title{Group-$k$ Consistent Measurement Set Maximization \\ for Robust Outlier Detection}


\author{Brendon Forsgren$^1$, Ram Vasudevan$^2$, Michael Kaess$^3$, Timothy W. McLain$^1$, Joshua G. Mangelson$^4$

\thanks{$^1$ Brendon Forsgren and Timothy McLain are with the Department of Mechanical Engineering, Brigham Young University {\tt\footnotesize bforsgren29@gmail.com, mclain@byu.edu}}
\thanks{$^2$ Ram Vasudevan is with the Department of Mechanical Engineering, University of Michigan {\tt\footnotesize ramv@umich.edu}}
\thanks{$^3$ Michael Kaess is with the Robotics Institute at Carnegie Mellon University {\tt\footnotesize kaess@cmu.edu}}
\thanks{$^4$ Joshua G. Mangelson is with the Department of Electrical and Computer Engineering, Brigham Young University {\tt\footnotesize mangelson@byu.edu}}
\thanks{This work has been funded by the Office of Naval Research under award number N00014-21-1-2435}
}

\maketitle

\begin{abstract}
    This paper presents a method for the robust selection of measurements
in a simultaneous localization and mapping (SLAM) framework. Existing
methods check consistency or compatibility on a pairwise basis, however many
measurement types are not sufficiently constrained in a pairwise scenario to
determine if either measurement is inconsistent with the other. This
paper presents group-$k$ consistency maximization (G$k$CM) that estimates
the largest set of measurements that is internally group-$k$ consistent.
Solving for the largest set of group-$k$ consistent measurements can be formulated as an
instance of the maximum clique problem on generalized graphs and can be solved by adapting current
methods. This paper evaluates the performance of G$k$CM using simulated data
and compares it to pairwise consistency maximization (PCM) presented in previous
work.

\end{abstract}

\section{Introduction}

In simultaneous localization and mapping (SLAM) a robot
collects data about its own trajectory and the environment.
Generating an accurate map requires that the robot
estimate its own trajectory while also estimating the locations of
environmental features being tracked.

SLAM is often represented as a factor graph with variable nodes including
pose and environmental features, and factor nodes. The problem
is formulated as the maximum likelihood estimate (MLE) of the robot
trajectory given the measurements made along that trajectory.
Assuming independence and additive Gaussian noise in the measurement and process
models, this becomes a nonlinear least squares problem that can be solved
quickly using available solvers
\cite{kaess2012isam2, grisetti2011g2o,agarwal2012ceres}.

However, nonlinear least squares is susceptible to outliers and reliably
determining accurate factors for non-odometric measurements is difficult.
Much work has been done to enhance robustness for pose-to-pose
loop closure measurements in the single-agent case
\cite{sunderhauf2012towards,agarwal2013robust,latif2013robust} and
more recently in the multi-agent scenario \cite{mangelson2018pairwise}. Other
work has focused on robustly selecting measurements of other types such as range
\cite{olson2005single}, visual features \cite{shi2021robin}, and
point clouds \cite{shi2021robin,lusk2021clipper}.

Classical approaches to outlier detection, such as rejection gating, classify
measurements as inliers or outliers.
Rather than attempt to classify measurements as
inliers or outliers, we find the largest consistent set of measurements. In
our prior work \cite{mangelson2018pairwise}, the problem was formulated as a
combinatorial optimization problem that seeks to find the largest set of
pairwise consistent measurements. It was shown that the optimization problem
can be transformed into an instance of the maximum clique problem where
available algorithms can often optimally solve moderately sized problems in real time.

In this work we generalize the methodology described in
\cite{mangelson2018pairwise} to scenarios where checking pairwise consistency
is not sufficient. We make the following contributions:
\begin{enumerate}
    \item We extend the notion of pairwise consistency maximization to group-$k$ consistency maximization and show that it can be solved by finding the maximum clique of a generalized graph with $k$-tuple edges.
    \item We generalize existing branch and bound/heuristic algorithms from graph theory \cite{pattabiraman2015fast} to efficiently search for the maximum clique of generalized graphs.
    \item We apply the work to a range-based SLAM scenario where a mobile vehicle is receiving range measurements to static beacons.
    \item We release a parallelized implementation of our proposed algorithm \url{https://bitbucket.org/jmangelson/gkcm/src/master/}.
\end{enumerate}


\section{Related Work}
\label{section:related_work}

Methods to identify sets of consistent measurements have received a great deal of
attention in the SLAM literature because, in many instances, a single inconsistent
measurement is enough to warp the estimated map. Much of the literature
focuses on identifying loop closures in pose graph SLAM where many methods set
high likelihood thresholds to remove false positives \cite{ozog2016long}. Other
methods like switchable constraints and dynamic covariance scaling
\cite{sunderhauf2012towards, agarwal2013robust} turn off measurements that have
a high residual error. Graduated non-convexity \cite{yang2020graduated} is an
approach that solves a convex approximation and iteratively
solves non-convex approximations of the original problem until the original problem is solved.
Max-mixtures \cite{olson2013inference} is a
technique that uses mixtures of Gaussians to model various data modes.
All these methods require an initialization and can fail with a poor
initial guess.

Random sample consensus (RANSAC) is a popular algorithm, especially in the
computer vision literature. The RANSAC algorithm
determines inlier/outlier sets by iteratively fitting models to random samples
of the data. The number of inliers is counted for each model and the model that
contains the highest number of inliers is selected \cite{andrew2001multiple}.
The process used to select the inlier set
makes the RANSAC algorithm perform poorly when the outlier ratio is large,
often resulting in a poor set of measurements being chosen. An improved
algorithm called RANSIC, was detailed in \cite{sun2021iron}
that utilizes a compatibility score among the random samples when
performing point cloud registration and achieves robustness against high
quantities of outliers.

Carlone et al. showed in \cite{carlone2014selecting} that determining if a
measurement is an inlier or outlier is unobservable and a better approach
is to find a set of internally coherent measurements and graph-based methods
have become popular in finding coherent sets.
One of the first such approaches by Bailey et al.
\cite{bailey2000data} proposed a maximum common subgraph algorithm to match
point clouds from a 2D scanning laser. Single-cluster graph partitioning (SCGP)
\cite{olson2005single} and CLIPPER \cite{lusk2021clipper} utilize spectral
relaxation to efficiently determine sets of consistent measurements.
PCM was introduced in \cite{mangelson2018pairwise} to select consistent inter-robot
loop closure measurements in multi-agent scenarios by solving an instance of
the maximum clique problem. Chang et al. \cite{chang2021kimera} introduced a heuristic that
decreased the run time of searching for the maximum clique when running PCM in
an incremental fashion. The notion of data similarity was combined with
pairwise consistency in \cite{do2020robust} to create a combined edge weight
that was used in solving a maximum edge weight clique problem to determine
the largest set of consistent measurements.

All the works mentioned in the previous paragraph check consistency on a pairwise basis. Shi
et al. \cite{shi2021robin} present ROBIN which generalizes the pairwise check
to group-$k$, where $k$ is the number of measurements required to check the
relevant invariant. Additionally, ROBIN approximates
the maximum clique by computing the \emph{maximum $k$-core} which is fast to compute
and often provides a good approximation to the maximum clique. However, this
approximation biases ROBIN towards accepting outlier measurements as opposed
to our approach which seeks to exclude all outliers.

We present G$k$CM, an algorithm that solves a similar problem to ROBIN in that
we extend the notion of consistency to a group-$k$ sense.
G$k$CM is different than ROBIN in that ROBIN aims to decrease the number of
outliers to a point where current solvers, such as RANSAC or GNC, work well.
Our work aims to
remove all measurements that are not suitable to include in a nonlinear least
squares problem without the need for other robust optimization methods.
We note that our approach is most similar to ROBIN but we evaluate G$k$CM
on a problem similar to that presented in \cite{olson2005single} used to test SGCP where
the consistency of range measurements from a moving vehicle to static beacons
is determined. Our work differs from \cite{olson2005single} in that SCGP
evaluates the consistency of range measurements on a pairwise basis while we
enforce consistency on a $k=4$ basis.

\section{Problem Formulation}
\label{section:problem_formulation}

In our factor graph formulation of range-only landmark SLAM, we denote the
discretized poses of the robot trajectory by $\bvec{x}_i \in$ SE(2) or SE(3)
and the positions of static beacons $\bvec{l}_k \in \mathbb{R}^3$. Factors in
the graph are derived from
the measurements observed by the robot and penalize estimates of the map and
trajectory that make observed measurements unlikely. We denote odometry
measurements that relate variables $\bvec{x}_i$ and $\bvec{x}_j$ by
$\bvec{z}_{ij}$. Likewise, we denote range measurements that relate variables
$\bvec{x}_i$ and $\bvec{l}_k$ as $\bvec{r}_{ik}$. The goal of SLAM is to
estimate the most likely value of each pose variable $\bvec{x}_i$ and beacon
variable $\bvec{l}_k$ given the measurements $\bvec{z}_{ij}$ and
$\bvec{r}_{ik}$. The problem can be formulated as the MLE problem
\begin{equation}
    \centering
    \hat{\mathbf{X}}, \hat{\mathbf{L}} = \underset{\bvec{X}, \bvec{L}}{\operatorname{argmax}} \ P(\bvec{Z}, \bvec{R}|\bvec{X}, \bvec{L})
    \label{eq:mle_problem}
\end{equation}
where, $\bvec{X}$ is the set of all pose variables $\bvec{x}_i$,
$\bvec{L}$ the set of all beacon variables $\bvec{l}_k$, $\bvec{Z}$ the set
of all odometry measurements $\bvec{z}_{ij}$, and $\bvec{R}$ the set of
all range measurements $\bvec{r}_{ik}$. Assuming there are outliers in
$\bvec{R}$, our goal becomes to select the largest set $\bvec{R}^* \subset
\bvec{R}$ that is internally consistent. Existing methods do this on the premise
that $\bvec{R}^*$ is pairwise consistent but we explore selecting
the set based on group consistency.

\section{Group-$k$ Consistency Maximization}
\label{sec:group_k_consist_max}

In this section we generalize the notion of consistency to sets of $k>2$
measurements and use this generalized definition to formulate
a combinatorial optimization problem.

While maximizing pairwise consistency in \cite{mangelson2018pairwise}
outperformed other existing robust SLAM methods, pairwise consistency is not
always a sufficient constraint to remove outlier measurements. For example, a set of range measurements may
all intersect in a pairwise manner even if the set of measurements do not
intersect at a common point indicating that they are pairwise consistent but
not group-3 consistent.

As currently framed, the consistency check described in
\cite{mangelson2018pairwise} is only dependent on two measurements. In some
scenarios, such as with the range measurements described above, we may want to
define a consistency function that depends on more than two measurements.

\subsection{Group-$k$ Consistency}

To handle the situation where consistency should be enforced in
groups of greater than two measurements we now define a novel notion of
\textit{group-$k$ internally consistent sets}.

\begin{definition}
  A set of measurements $\bvec{\widetilde{Z}}$ is \textbf{group-$k$ internally consistent} with respect to a consistency metric $C$ and the threshold $\gamma$ if
\begin{align}
  C(\{\bvec{z}_o, \cdots, \bvec{z}_k\}) \leq \gamma, \quad \forall \quad \{\bvec{z}_0, \cdots, \bvec{z}_k\} \in \mathcal{P}_k(\bvec{\widetilde{Z}})
  \label{eq:gk_consistency}
\end{align}
where, $C$ is a function measuring the consistency of the set of measurements $\{\bvec{z}_o, \cdots, \bvec{z}_k\}$, $\mathcal{P}_k(\mathbf{\bvec{\widetilde{Z}}})$ is the the set of all
permutations of $\bvec{\widetilde{Z}}$ with cardinality $k$, and $\gamma$ is chosen a priori.
\end{definition}
This definition of consistency requires that every combination of measurements
of size $k$ be consistent with $C$ and $\gamma$.
The appropriate choice of consistency function is problem dependent and
therefore left to the user to determine, however we define our consistency
function for our range-based SLAM problem in \cref{sec:range_slam}.

As with pairwise consistency, establishing group-$k$ consistency does not
guarantee full joint consistency. We settle for checking group-$k$ consistency
and use it as an approximation for joint consistency to keep the problem tractable.

\subsection{Group-$k$ Consistency Maximization}

Analagous to pairwise consistency maximization defined in
\cite{mangelson2018pairwise}, we now want to find the largest subset of
measurements that is internally \emph{group-$k$ consistent}. The following
assumptions are used by our method:
\begin{assumption}
    Data association for the range measurements is known
    (i.e. measurements to different beacons are known to be inconsistent). We
    will relax this assumption later in one of our experiments.
\end{assumption}
\begin{assumption}
    The system used to derive range measurements is not biased toward selecting
    incorrect measurements over correct ones.
    \label{assump:correct_meas}
\end{assumption}
If the above assumptions, especially \cref{assump:correct_meas}, are
true then we can make the following assumption.
\begin{assumption}
    As the number of range measurements increases, the number of measurements
    in the correct consistent subset will grow larger than those in other
    subsets.
\end{assumption}
As in PCM, our goal is to find the largest consistent subset of $\bvec{Z}$.
We accomplish this by introducing a binary
switch variable $s_u$ for each measurement in $\bvec{Z}$ and let $s_u$ be 1 if the
measurement is contained in the chosen subset and 0 otherwise. Letting $\bvec{S}$ be
the vector containing all $s_u$, our goal is to find $\bvec{S}^*$ to the following
optimization problem
\begin{equation}
  \begin{gathered}
    \bvec{S}^{*} = \underset{\bvec{S}\in\{0,1\}^m}{\operatorname{argmax}} \: \norm{\bvec{S}}_0 \\
    \text{s.t.}~~
 C(\{\bvec{z}_0, \cdots, \bvec{z}_k\}) ~ s_0 \cdots s_k \leq \gamma \\~~~~~~~~~~~~~~~~~~~~~~~ \forall \{\bvec{z}_0, \cdots, \bvec{z}_k\} \in \mathcal{P}_k(\bvec{Z})
\end{gathered}
\label{eq:comb_formulation_gkcm}
\end{equation}
where $m$ is the number of measurements in $\bvec{Z}$ and $\bvec{z}_u$ is the measurement
corresponding to $s_u$. We refer to this problem as the Group-$k$ Consistency
Maximization, or G$k$CM, problem. This problem is a generalization of PCM and
for $k=2$ they become identical.

\begin{figure}[!tb]%
  \centering%
  \subfigure[Generalized Graph]{%
    \includegraphics[width=0.4\columnwidth]{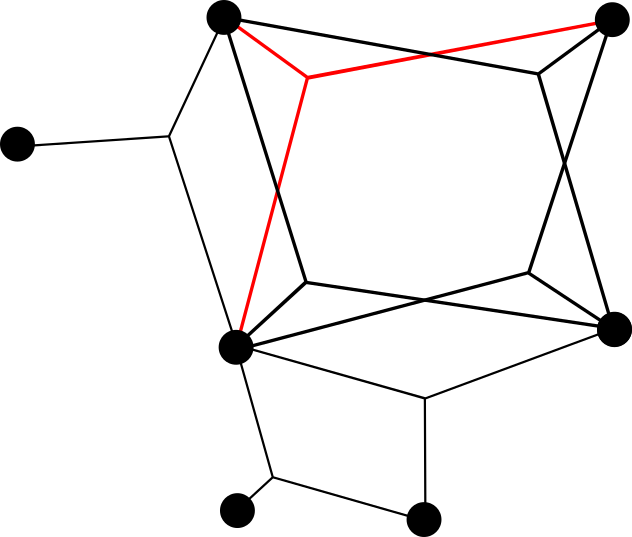}%
    \label{sfig:gen_graph_edge}%
  }%
  \hfil%
  \subfigure[Maximum Clique]{%
    \includegraphics[width=0.4\columnwidth]{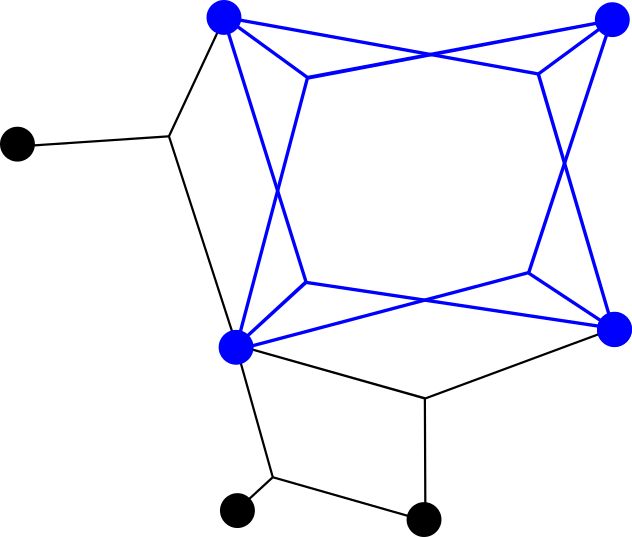}%
    \label{sfig:gen_graph_clique}%
  }%
  \caption{An example of a generalized consistency graph with edges made of $3$-tuples. \subref{sfig:gen_graph_edge} highlights that each edge denotes consistency of $3$ measurements. \subref{sfig:gen_graph_clique} highlights the maximum clique of the generalized graph in blue.}
  \label{fig:gen_graph}
\end{figure}

\subsection{Solving Group-$k$ Consistency Maximization}

As with PCM, we can solve the G$k$CM problem by finding the maximum clique of a
consistency graph. However, because we want to find the largest subset that is
group-$k$ internally consistent we need to operate over generalized graphs.
In graph theory, a \emph{k-uniform hypergraph} (or
\emph{generalized graph}), $G$, is defined as a set of vertices $V$ and a set
of $k$-tuples of those vertices $\mathcal{E}$ \cite{bollobas1965generalized}.
Each $k$-tuple is referred to as an edge and a \emph{clique} within this
context is a subgraph of $G$ where every possible edge is an edge in
$\mathcal{E}$. We now introduce the concept of a
\emph{generalized consistency graph}:
\begin{definition}
  A \textbf{generalized consistency graph} is a generalized graph
  $G={V, \mathcal{E}}$ with $k$-tuple edges, where each vertex
  $v \in V$ represents a measurement and each edge $e \in \mathcal{E}$
  denotes consistency of the vertices it connects.
\end{definition}
Solving \cref{eq:comb_formulation_gkcm} is equivalent to finding
the maximum clique of a generalized consistency graph and consists of the
following two steps:
\begin{enumerate}
  \item Building the generalized consistency graph
  \item Finding the maximum clique
\end{enumerate}
The next two sections explain these processes in more detail.

\section{Building the Generalized Consistency Graph}
\label{sec:build_graph}
The graph is built by creating a vertex for each measurement and
performing the relevant consistency checks to determine what edges should be
added. If the graph is created all at once, their are
$\binom{m}{k}$ checks to perform. If the graph is being
built incrementally by checking the consistency of a new added measurement with
those already in the graph then the number of checks is $\binom{m-1}{k-1}$.
This means that as $k$ increases the number of checks that need to be
performed increases factorially with $k$. Thus it is important that the
consistency function in \cref{eq:gk_consistency} be computationally efficient.
Note that all the checks are independent allowing for the computation to be
parallelized on a CPU or GPU to decrease the time to perform the necessary
checks.



\begin{algorithm*}[!tb]
    \small
  \caption{Exact Algorithm for Finding the Maximum Clique of a k-Uniform Hypergraph. \newline \textbf{Input:} Graph $G=(V,\mathcal{E})$,  \textbf{Output:} Maximum Clique $S_{max}$}
  \vspace{-5mm}
  \label{alg:gmc_exact}
  \begin{multicols}{2}
  \begin{algorithmic}[1]
    \Function{MaxClique}{$G=(V,\mathcal{E})$}
    \State $S_{max} \gets \emptyset$
    \For{$i=1$ to $n$} \label{alg:gmc_exact:node_loop}
    \If{$d(v_i)+1 \geq |S_{max}|$}
    \For{\textbf{each} $e \in E(v_i)$} \label{alg:gmc_exact:edge_for}
    \State $S \gets e \cup v_i; $ $U \gets \emptyset$
    \State $R \gets $ \Call{CombinationsOfSize}{$S, k-1$} \label{alg:gmc_exact:R}
    \For{\textbf{each} $v_j \in N(v_i)$}
    \If{$j > i$}
    \If{$d(v_j)+1 \geq |S_{max}|$}
    \If{$R \subset E(v_j)$} \label{alg:gmc_exact:limit_R}
    \State $U \gets U \cup \{v_j\}$    \label{alg:gmc_exact:U_init}
    \EndIf
    \EndIf
    \EndIf
    \EndFor
    \State \Call{Clique}{$G, R, S, U$}
    \EndFor
    \EndIf
    \EndFor
    \EndFunction
  \item[]
  \item[]
  \item[]
  \item[]
  \item[]
  \end{algorithmic}
  \begin{algorithmic}[1]
    \Function{Clique}{$G=(V,\mathcal{E})$, $R$, $S$, $U$}
    \If{$U = \emptyset$}
    \If{$|S| > |S_{max}|$}
    \State $S_{max} \gets S$ \label{alg:gmc_exact:update}
    \EndIf
    \EndIf
    \While{$|U| > 0$}
    \If{$|S| + |U| \leq |S_{max}|$}
    \State \textbf{return}
    \EndIf
    \State Select any vertex $u$ from $U$
    \State $U \gets U \setminus \{u\}; ~S_{rec} \gets S \cup \{u\}$ \label{alg:gmc_exact:update_S}
    \State $N'(u) := \{w | w \in N(u) $ \textbf{and} $d(w) \geq |S_{max}|\}$
    \State $U_{rec} \gets \emptyset;~R_{rec} \gets R$
    \For{\textbf{each} $ p\in $ \Call{CombinationsOfSize}{$S, k-2$}}
    \State $R_{rec} \gets R_{rec} \cup \{p \cup \{u\}\}$ \label{alg:gmc_exact:update_R}
    \EndFor
    \For{\textbf{each} $ q \in U \cap N'(u)$}
    \If{$R_{rec} \subset E(q)$}
    \State $U_{rec} \gets U_{rec} \cup \{q\}$ \label{alg:gmc_exact:update_U}
    \EndIf
    \EndFor
    \State \Call{Clique}{$G, R_{rec}, S_{rec}, U_{rec}$}
    \EndWhile
    \EndFunction
  \end{algorithmic}
  \end{multicols}
  \vspace{-5mm}
\end{algorithm*}

\section{Finding the Maximum Clique of a Generalized Graph}
\label{sec:max_cliqu_of_generalized_graph}

Once the graph has been built, we can find the largest consistent set by
finding the maximum clique of the graph. The PCM algorithm used the exact and
heuristic methods presented by \cite{pattabiraman2015fast} but these algorithms
were not designed for generalized graphs and used only a single thread. Here we
generalize their algorithms to $k$-uniform hypergraphs and provide a
parallelized implementation of their algorithms.

\begin{figure}[!tb]%
  \centering%
  \includegraphics[width=.93\columnwidth]{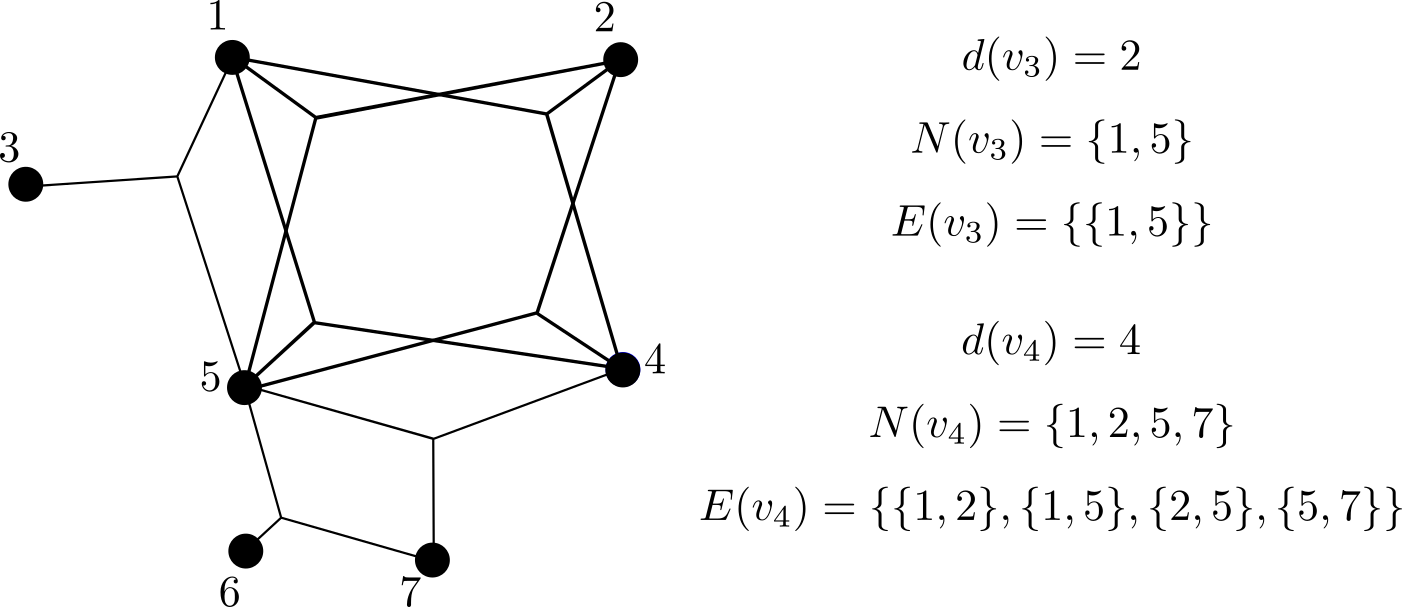}%
  \caption{Examples of the degree, neighborhood, and edge set definitions for generalized graphs.}%
  \label{fig:gen_graph_notation}%
\end{figure}

\begin{algorithm*}
    \small
  \caption{Heuristic Algorithm for Finding the Maximum Clique of a k-Uniform Hypergraph. \newline \textbf{Input:} Graph $G=(V,\mathcal{E})$,  \textbf{Output:} Potential Maximum Clique $S_{max}$}
  \vspace{-5mm}
  \label{alg:gmc_heuristic}
  \begin{multicols}{2}
  \begin{algorithmic}[1]
    \Function{MaxCliqueHeu}{$G=(V,\mathcal{E})$}
    \State $S_{max} \gets \emptyset$
    \For{$i=1$ to $n$}
    \If{$d(v_i)+1 \geq |S_{max}|$} \label{alg:gmc_heuristic:degree_check}
    \State Select $e \in E(v_i)$ with max connect. in $E(v_i)$ \label{alg:gmc_heuristic:select_edge}
    \State $S \gets e \cup v_i; $ $U \gets \emptyset$
    \State $R \gets $ \Call{CombinationsOfSize}{$S, k-1$}
    \For{\textbf{each} $v_j \in N(v_i)$}
    \If{$d(v_j)+1 \geq |S_{max}|$}
    \If{$R \subset E(v_j)$}
    \State $U \gets U \cup \{v_j\}$
    \EndIf
    \EndIf
    \EndFor
    \If{$|S| + |U| > |S_{max}|$}
    \State \Call{CliqueHeu}{$G, R, S, U$}
    \EndIf
    \EndIf
    \EndFor
    \EndFunction
    \item[]
  \end{algorithmic}
  \begin{algorithmic}[1]
    \Function{CliqueHeu}{$G=(V,\mathcal{E})$, $R$, $S$, $U$}
    \If{$U = \emptyset$}
    \If{$|S| > |S_{max}|$}
    \State $S_{max} \gets S$
    \EndIf
    \EndIf
    \State Select a vertex $u \in U$ with max connect. in $E(v_i)$ \label{alg:gmc_heuristic:select_vertex}
    \State $U \gets U \setminus \{u\}; ~S_{rec} \gets S \cup \{u\}$
    \State $N'(u) := \{w | w \in N(u) $ \textbf{and} $d(w) \geq |S_{max}|\}$
    \State $U_{rec} \gets \emptyset;~R_{rec} \gets R$
    \For{\textbf{each} $ p\in $ \Call{CombinationsOfSize}{$S, k-2$}}
    \State $R_{rec} \gets R_{rec} \cup \{p \cup \{u\}\}$
    \EndFor
    \For{\textbf{each} $ q \in U \cap N'(u)$}
    \If{$R_{rec} \subset E(q)$}
    \State $U_{rec} \gets U_{rec} \cup \{q\}$
    \EndIf
    \EndFor
    \State \Call{CliqueHeu}{$G, R_{rec}, S_{rec}, U_{rec}$}
    \EndFunction
  \end{algorithmic}
  \end{multicols}
  \vspace{-4mm}
\end{algorithm*}

We start by defining relevant notation. We denote the $n$ vertices of the graph
$G=(V, \mathcal{E})$ as $\{v_1, \cdots, v_n\}$. Each vertex has a neighborhood $N(v_i)$,
that is the set of vertices connected to that vertex by at least one edge. The
degree of $v_i$, $d(v_i)$, is the number of vertices in its neighborhood.
We also define an edge set, $E(v_i)$, for each vertex consisting of a set of
$(k-1)$-tuples of vertices. The edge set is derived from the set of $k$-tuples
in $\mathcal{E}$ containing the given vertex by removing the given vertex from
each edge. \Cref{fig:gen_graph_notation} shows an example of these values for a
given graph.

\subsection{Algorithm Overview}
The generalized exact and heuristic algorithms presented in
\cref{alg:gmc_exact} and \cref{alg:gmc_heuristic} respectively are similar in
structure to the algorithms in \cite{pattabiraman2015fast} but require
additional checks to guarantee a valid clique is found since the algorithms now
operate over generalized graphs.

The exact algorithm, \cref{alg:gmc_exact}, begins with a vertex $v$ and finds
cliques of size $k$ that contain $v$ (MaxClique line
\ref{alg:gmc_exact:edge_for}). A set of vertices, $U$, that would increase the
clique size by one is found (MaxClique line \ref{alg:gmc_exact:limit_R}) from the set of edges $R$
that a valid candidate vertex must have (MaxClique line \ref{alg:gmc_exact:R}).
The Clique function then recursively iterates through potential cliques and
updates $R$ and $U$ (Clique lines \ref{alg:gmc_exact:update_R},
\ref{alg:gmc_exact:update_U}). The clique is tracked with $S$ and a check is
performed to see if $S>S_{max}$ where $S_{max}$ is replaced with $S$ if the
check passes. The process is repeated for each vertex in the graph
(MaxClique line \ref{alg:gmc_exact:node_loop}). The exact algorithm
evaluates all possible cliques and as such, the time complexity of the exact
algorithm is exponential in worse case.

The heuristic algorithm, \cref{alg:gmc_heuristic}, has a similar structure to
the exact algorithm but uses a greedy search to find a potential maximum clique
more quickly. For each node with a degree greater than the size of the current
maximum clique (MaxCliqueHeu line \ref{alg:gmc_heuristic:degree_check}) the
algorithm selects a clique of size $k$ who has the greatest number of
connections in $E(v_i)$ (MaxCliqueHeu line
\ref{alg:gmc_heuristic:select_edge}). This is done by summing the number of
connections each
node in $N(v_i)$ has in $E(v_i)$ and selecting the edge $e \in E(v_i)$ with the
sum total of connections. It the selected clique can potentially be made larger
than $S_{max}$ then a greedy search selects nodes based on the largest number
of connections in $E(v_i)$ (CliqueHeu line
\ref{alg:gmc_heuristic:select_vertex}). The heruistic algorithm presented in
\cref{alg:gmc_heuristic} has the same complexity of $O(n\Delta^2)$ presented in
\cite{pattabiraman2015fast} despite the modifications made to operate on
generalized graphs.

Both algorithms are gauranteed to find a valid clique and can be easily parallelized by using multiple threads to
simultaneously evaluate each iteration of the for loop on line
\ref{alg:gmc_exact:node_loop} of MaxClique and MaxCliqueHeu. This significantly
decreases the run-time of the algorithm. Our released C++ implementation allows
the user to specify the number of threads to be used.

\subsection{Evaluation of $\operatorname{MaxClique}$ and $\operatorname{MaxCliqueHeu}$}

We carried out two experiments to evaluate the effectiveness of
\cref{alg:gmc_exact} and \cref{alg:gmc_heuristic}.

\subsubsection{Timing Comparison} 
\begin{figure}[!tb]%
  \centering%
  \includegraphics[width=.93\columnwidth]{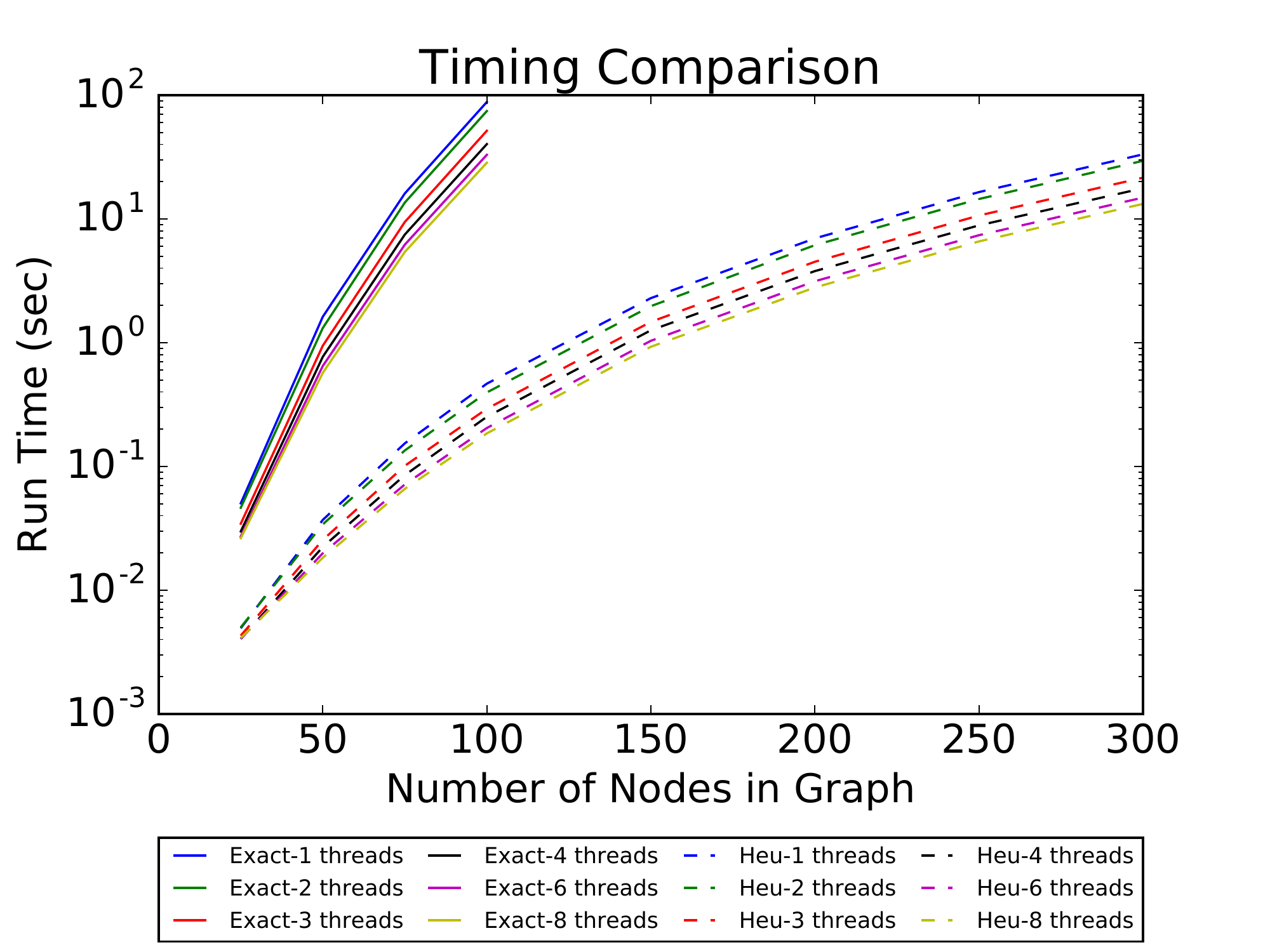}%
  \caption{Average run-time for both the exact and heuristic generalized maximum clique algorithms proposed
    in this paper. This includes both the time to evaluate the necessary
    data-structures such as neighborhoods/edge sets and the time to estimate the maximum clique.
    Using eight threads, the heuristic algorithm was able to find the maximum clique of a graph
    with 250 nodes in a few seconds. }%
  \label{fig:gmc_timing}%
\end{figure}

In the first experiment, we randomly
generated 3-uniform hypergraphs with various node counts ranging from 25
vertices to 300 vertices. Each graph contained all the edges necessary
to contain a maximum clique of cardinality 10 and additional randomly selected
edges to meet a specified graph density. While the run-time of the algorithm is
dependent on the density of the graph, for this
experiment, we chose to hold the density of the graph constant at 0.1 such
that approximately 10 percent of all potential edges were contained in the
graph. We generated 100 sample graphs for each
number of nodes. We then used both the exact and heuristic algorithm to
estimate the maximum clique of each
graph and measured the average run-time for each.  \Cref{fig:gmc_timing}
shows the results of this experiment
using various numbers of threads ranging from one to eight. The exact algorithm
was only used for graphs with
a total number of nodes of 100 or less because of the exponential nature of the
algorithm.

\subsubsection{Heuristic Evaluation} 
\begin{figure}[!tb]%
  \centering%
  \includegraphics[width=.93\columnwidth]{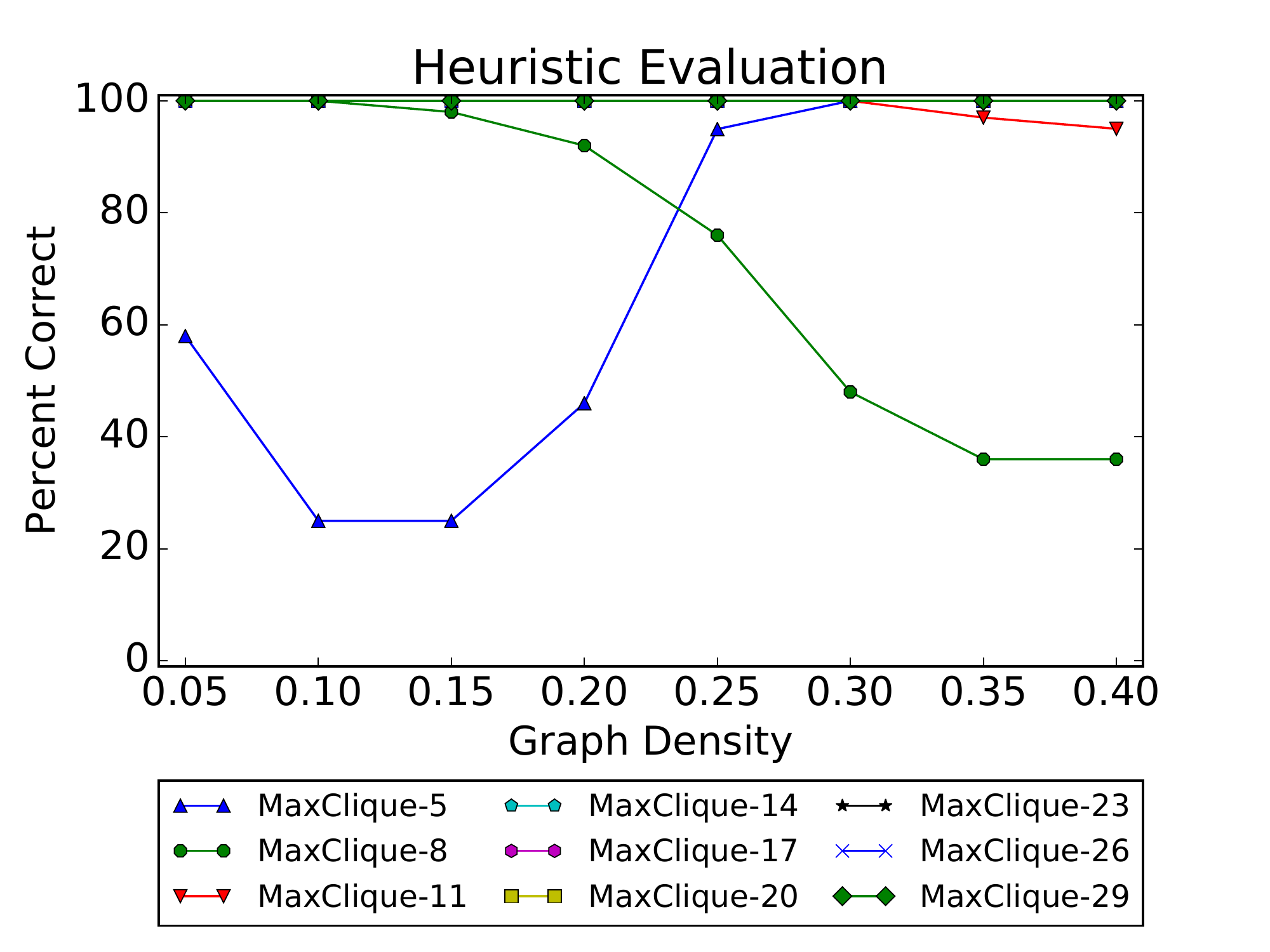}%
  \caption{Evaluation of the heuristic algorithm proposed in Algorithm \ref{alg:gmc_heuristic}. Individual
    lines denote the cardinality of the maximum clique inserted into the graph. The horizontal axis denotes the density of
    edges in the graph and the vertical axis denotes the percentage of test cases where the algorithm returned a
    clique of the correct cardinality. The heuristic algorithm returned cliques of the correct size 100 percent
    of the time for the graphs with max clique size of 14, 17, 20, 23, 26, and 29. }%
  \label{fig:gmc_heuristic_evaluation}%
\end{figure}

In the second experiment, we again
randomly generated 3-uniform hypergraphs,
however, in this case we varied the density of the graph and the size of the
inserted clique, while holding the
total number of nodes at 100. For each graph we used the MaxCliqueHeu algorithm
to estimate the
maximum clique and then evaluated whether or not the algorithm was successful
in finding a clique of the same
size as the clique we inserted. We again generated 100 sample graphs for each
combination of inserted-clique
size and graph density. \Cref{fig:gmc_heuristic_evaluation} plots the
summarized results. If the algorithm
happened to return a maximum-clique larger, then the inserted clique than the
associated sample was dropped.

This experiment shows that the
size of the maximum clique and the success rate of the proposed heuristic
algorithm are correlated. In addition,
it shows that with the exception of the case when the inserted clique was very
small (cardinality 5), the density of the
graph and the success rate are inversely correlated. As such,
the heuristic seems to perform best when the size of the maximum clique is
large and/or when the connectivity of the
graph is relatively sparse.


\section{Range-based SLAM}
\label{sec:range_slam}
For the remainder of this paper we will consider G$k$CM in the context of
a range-based SLAM scenario and will use the following $k=4$ consistency check,
\begin{equation}
    \centering
    \begin{gathered}
    C(\bvec{r}_{ai}, \bvec{r}_{bi}, \bvec{r}_{ci}, \bvec{r}_{di}) =
    \norm{h(\bvec{X}_{abcd}, \bvec{R}_{abc}^i) - \bvec{r}_{di}}_{\Sigma} \leq \gamma
    \end{gathered}
    \label{eq:range_consistency_check}
\end{equation}
where $\bvec{r}_{di}$ is a range measurement from pose $d$ to beacon $i$,
$\bvec{X}_{abcd}$ is a tuple of poses $a$, $b$, $c$, and $d$, and $\bvec{R}_{abc}^i$
is a tuple of range measurements from poses $a$, $b$, and $c$ to beacon $i$.
The value $\gamma$ is a threshold value and the function
$h(\bvec{X}_{abcd}, \bvec{R}_{abc}^i)$ is a measurement model defined as
\begin{equation}
    \centering
    h(\bvec{X}_{abcd}, \bvec{R}_{abc}^i) = \norm{\bvec{l}(\bvec{X}_{abc}, \bvec{R}_{abc}^i) - \bvec{p}_{d}}_2
\end{equation}
where $\bvec{X}_{abc}$ is a tuple of poses $a$, $b$, and $c$, and
$\bvec{p}_i$ is the position of pose $i$.
The function $ \bvec{l}(\bvec{X}_{abc}, \bvec{R}_{abc}^i)$ is a
trilateration function that depends on the poses and the range measurements
received at poses $a$, $b$, and $c$ and returns an estimate of the beacon's
location. The covariance, $\Sigma$, is a function of the covariances on the
measurements $\bvec{r}$ and the poses $\bvec{x}$. The joint covariance,
$\Sigma_j$, of the poses
and beacon location are calculated by forming the measurement Jacobian of a
factor graph and using methods described in \cite{kaess2009covariance}. Once
the joint covariance has been obtained the covariance is calculated as
$\Sigma = H \Sigma_T H^T$ where
$H = \frac{\partial h} {\partial \bvec{x}, \bvec{l}, \bvec{r}_d}$ and
$\Sigma_T = \mathit{blockdiag}(\Sigma_j, \Sigma_{r_d})$.

The metric checks that the range to the intersection point of
three range measurements matches the range of the fourth
measurement. The check is done four times for a given set of four measurements
where each permutation of three measurements is used to localize the beacon.
Given the combinatorial nature of the number of checks
to be performed, the trilateration algorithm needs to be fast and accurate.
The algorithm described in \cite{zhou2011closed} fits these criteria and
presents a closed form algorithm that performs comparably to an iterative
nonlinear optimization approach but without the need for an initial guess
or an iterative solver.

\subsection{Degenerate Configurations}
\begin{figure}[!tb]
    \centering
    \includegraphics[width=0.93\columnwidth]{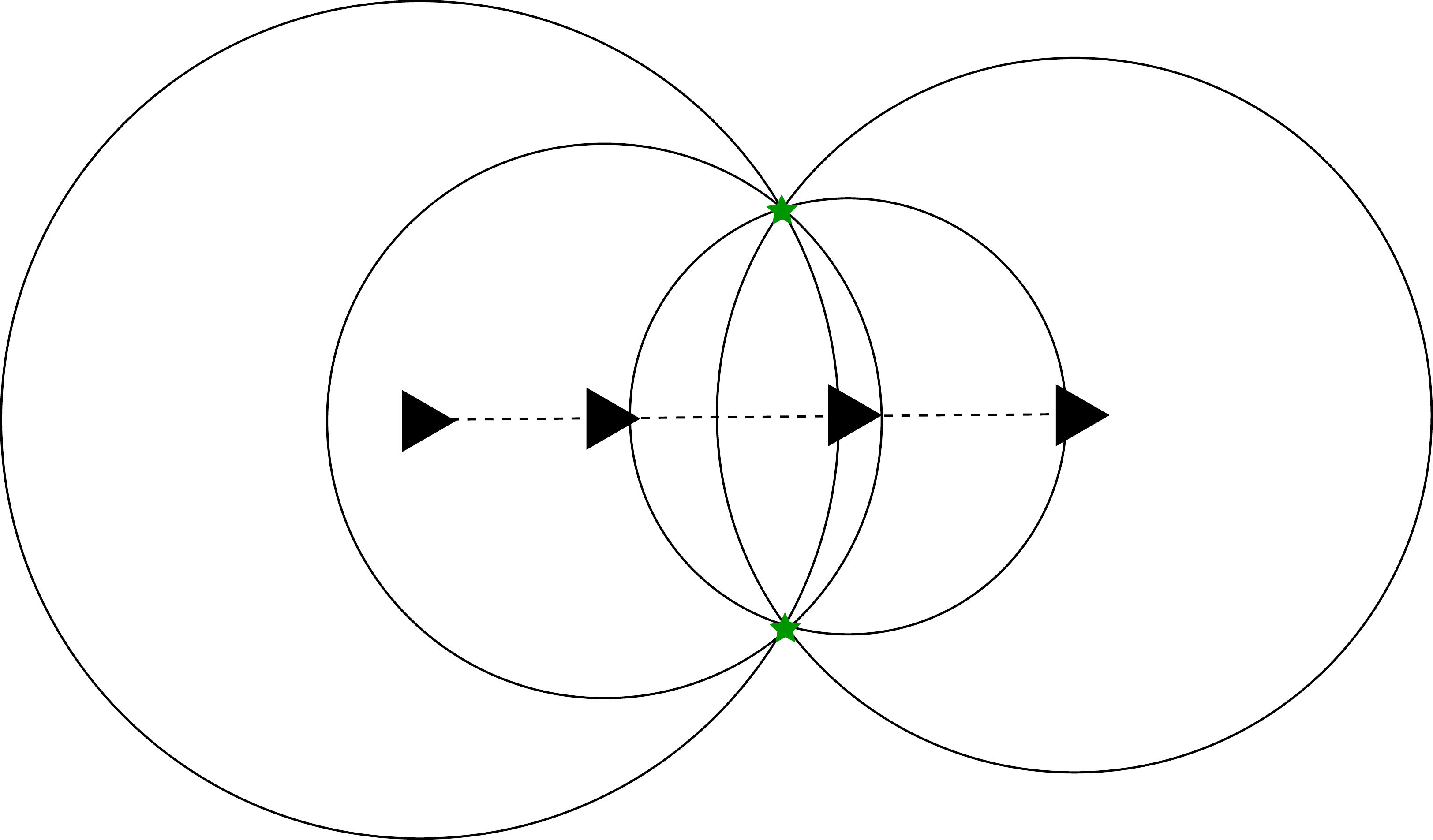}
    \caption{Degenerate pose configuration where range measurements do not result in a unique landmark location.}
    \label{fig:degenerate_config}
\end{figure}

Since our consistency check defined in \cref{eq:range_consistency_check} uses
a trilateration algorithm we need to discuss the scenarios
where trilateration fails to provide a unique solution.
The first case is where the poses are collinear as shown in
\cref{fig:degenerate_config} and the second is when two of the three
poses occupy the same position. The trilateration algorithm in
\cite{zhou2011closed} can return two estimates for the beacon's
location and the consistency check can pass if either estimate is deemed
consistent.

If such a solution is not available, then a test to detect a degeneracy
can be designed. If the test indicates the poses are in a
degenerate configuration, the scenario can be handled by storing one or more of
the measurements in a buffer whose consistency with the maximum clique can be
tested after the clique has been found. If a degeneration is still present then
the consistency of the measurement must be tested another way or the
measurement be labeled inconsistent. In practice, we found that degenerate
configurations did not present an issue because the odometry noise caused pose
estimates used in the consistency check to not be degenerate
even when the true configuration of poses was degenerate.

\section{Results}
\label{section:results}

In this section we evaluate the performance of G$k$CM on several synthetic
datasets where a robot is exploring and taking range measurements to static
beacons. We compare the results of G$k$CM to the results of PCM where the
consistency check for PCM is the check used in \cite{olson2005single}. Due to
runtime constraints all results are presented using the heuristic algorithm
presented in \cref{alg:gmc_heuristic}.

\subsection{Simulated 2D World}
First we simulate a two-dimensional world where a robot is navigating in the
plane. We simulate three different trajectories, (Manhattan world, circular, and a
straight line) along with range measurements to static beacons placed randomly
in the world. Gaussian noise was added to all range measurements and a portion
of the measurements were corrupted to simulate outlier measurements. Half of
the corrupted measurements were generated in clusters of size 5 and the other
half as single random measurements
using a Gaussian distribution with a random mean and a known variance.
We assume that the variances of the range measurements are known
and that these variances are used when performing the
consistency check. The simulation was run multiple times varying values such as
the trajectory and beacon locations, and statistics were recorded to compare
G$k$CM with PCM.

\subsubsection{Monte Carlo Experiment}
\begin{figure}[!tbh]
    \vspace{-3mm}
    \centering
    \includegraphics[width=0.93\columnwidth, trim=0 0 0 25, clip]{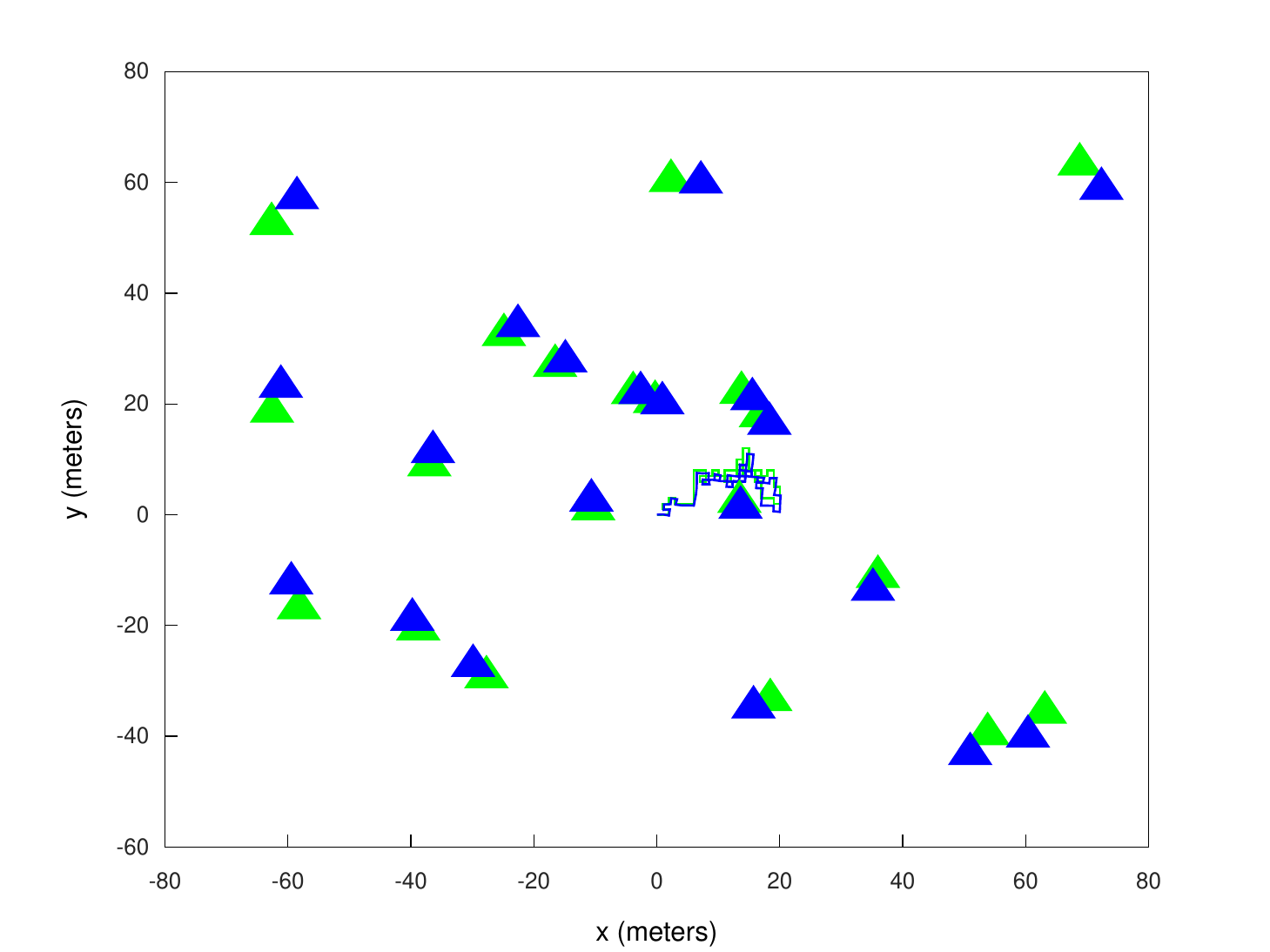}
    \caption{A larger experiment where 80 percent of the measurements are outliers. The green line and triangles are the true trajectory and beacon locations respectively while blue are the estimated trajectory and locations.}
    \label{fig::large_mc}
\end{figure}

This first example was done to show how well G$k$CM
performs in situations with large percentages of outliers.
In this experiment a trajectory of 100 poses was simulated with measurements
being taken at each pose and 80 of the measurements were corrupted to be
outliers. G$k$CM was used to identify consistent measurements which were
used used to solve the range-based SLAM problem in \cref{eq:mle_problem}
using GTSAM \cite{kaess2012isam2}.
The experiment averaged statistics over 81 runs and results are shown in
\cref{tab:large_mc_stats}. G$k$CM outperforms PCM in every
metric except the number of inliers found. Since to goal is to reject outliers,
excluding a certain number of inliers is acceptable as long as outliers are
also excluded.
We primarily use the true positive rate (TPR), false positive
rate (FPR) and $\chi^2$ value to evaluate how well how G$k$CM and PCM perform.
Ideal values for these statistics are respectively 1, 0 and $\chi^2 < 3.84$
indicating the estimates fit the measurements with 95\% confidence.
Additionally, we show the median $\chi^2$ value. The large
difference between the mean and the median, as well as the large standard
deviation indicate that G$k$CM is performs better than the mean indicates.
Looking at the $\chi^2$ values from all runs shows that the mean is greater
than 75\% of all the values showing that the times when G$k$CM failed skewed
the mean.
\Cref{fig::large_mc} shows a sample map output by GTSAM when
using the set of measurements selected by G$k$CM.

\begin{table*}[!tbh]
    \centering
    \vspace{3mm}
    \caption{Statistics for G$k$CM and PCM in Monte Carlo experiment. Best results are in \textbf{BOLD}}

\scalebox{1.0}[1.0]{
\begin{tabular}{|c|cc|cc|cc|cc|cc|ccc|}
\hline
\multirow{2}{*}{} & \multicolumn{2}{c|}{Trans. RMSE (m)}       & \multicolumn{2}{c|}{Rot.  RMSE (rad)}     & \multicolumn{2}{c|}{Beacon Error (m)}      & \multicolumn{2}{c|}{Residual}             & \multicolumn{2}{c|}{Inliers}              & \multicolumn{3}{c|}{$\chi^2$}                                       \\ \cline{2-14}
                  & \multicolumn{1}{c|}{Avg} & Std             & \multicolumn{1}{c|}{Avg} & Std            & \multicolumn{1}{c|}{Avg} & Std             & \multicolumn{1}{c|}{Avg} & Std            & \multicolumn{1}{c|}{TPR} & FPR             & \multicolumn{1}{c|}{Avg} & \multicolumn{1}{c|}{Std} & Median        \\ \hline
G$k$CM            & \textbf{1.9774}          & \textbf{1.8509} & \textbf{0.2805}          & \textbf{0.077} & \textbf{12.118}          & \textbf{28.114} & \textbf{442.82}          & \textbf{948.3} & 0.85                     & \textbf{0.007} & \textbf{2.28}            & \textbf{4.94}            & \textbf{0.56} \\ \hline
PCM               & 7.8664                   & 8.3322          & 0.5767                   & 0.2405         & 26.883                   & 43.502          & 18460                    & 23355          & \textbf{0.92}            & 0.026          & 89.68                    & 111.26                   & 52.69         \\ \hline
\end{tabular}
}
    \label{tab:large_mc_stats}
    \vspace{-4mm} 
\end{table*}

\begin{figure}[!tbh]
\vspace{-3mm} 
    \centering
    \includegraphics[width=0.93\columnwidth, trim=0 0 0 25, clip]{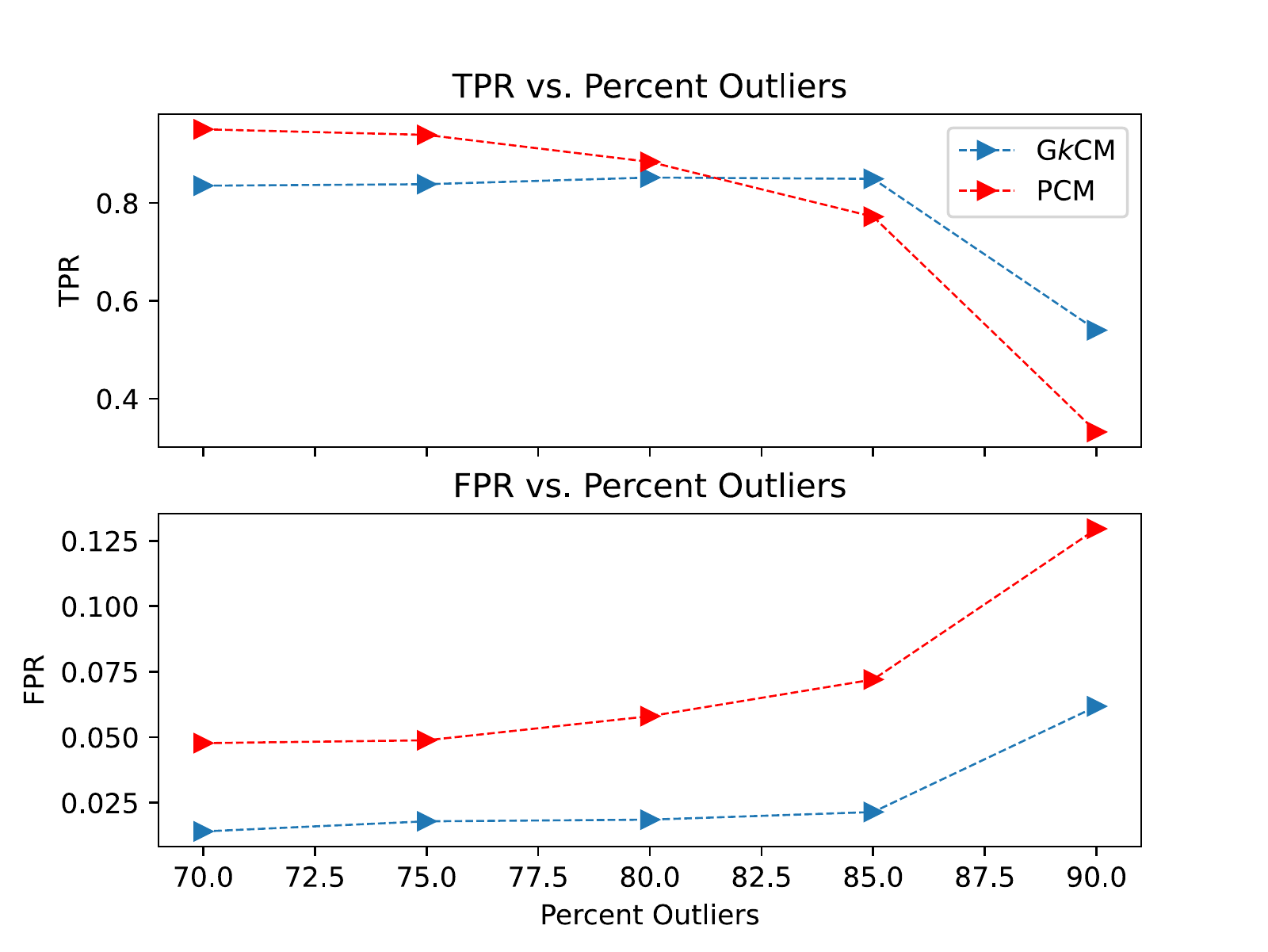}
    \caption{Results showing the normalized TPR ($\mathit{TP}/(\mathit{TP} + \mathit{FN}))$ and FPR $\mathit{FP}/(\mathit{FP} + \mathit{TN})$ by varying the number of outliers for a fixed trajectory.}
    \label{fig:varied_outlier_ratio}
\end{figure}

Additionally, we wished to know at what ratio of outliers to inliers did the
performance of G$k$CM begin to drop off. To measure this we simulated robot
odometry for 100 poses and corrupted the measurements taken to a beacon with  enough
outliers to achieve a certain percentage of outliers. We ran the set of
measurements through G$k$CM and observed if the selected set of consistent
measurements matched the set of inlier measurements. Using the same robot
odometry, this was done with several different outlier percentages. The
process was repeated for multiple trajectories and the true/false positive
rates for each outlier percentage were recorded. Results can be seen in
\cref{fig:varied_outlier_ratio}.

The figure shows that the true and false positive rates for
G$k$CM are fairly constant until about 85 percent of the measurements are
outliers while the true positive rate decreases with the number of outliers for
PCM and the false positive rate increases. These results are expected because
as more outliers are present, it is more likely that either an outlier clique
will form or that an outlier measurement will intersect with the inlier set with
a pairwise basis than a group-4 basis. Thus showing the need for group
consistency.

\subsection{Data Association}
In this experiment we remove the assumption that
the correspondence between a range measurement and its
beacon is known. To accomplish this, we modified both the exact and heuristic
algorithms in order to track the $n$ largest
cliques where $n$ is the number of beacons in the environment assuming the
number of beacons is known. Since each clique corresponds
to consistent measurements that belong to a unique beacon, we enforce the
constraint that a measurement cannot appear in more than one clique.

\begin{figure}[!tbh]
\vspace{-4mm} 
    \centering
    \includegraphics[width=0.93\columnwidth, trim=0 0 0 25]{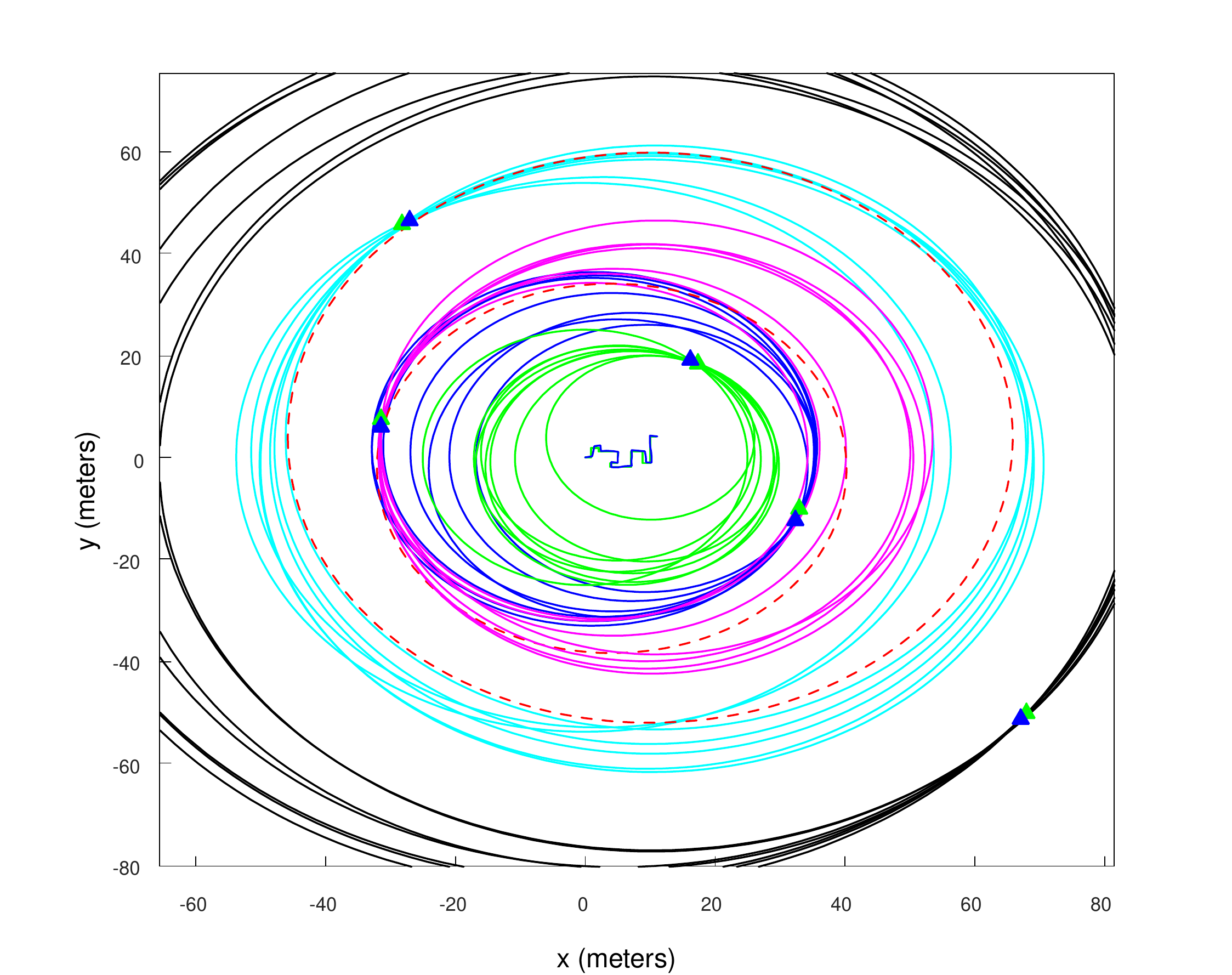}
    \caption{Results of G$k$CM for performing data association and outlier rejection. Each clique found is shown in a different color. Measurements labeled as outliers included in the maximum clique are red dashed lines.}
    \label{fig:data_association}
\end{figure}

This experiment was run on a short trajectory of 30 poses
where five measurements were received at each pose (one to each beacon).
As such there are 150 measurements being considered by the G$k$CM
algorithm. Results were averaged over 81 different trials.
Visual results can be seen in
\cref{fig:data_association} while statistics are in
\cref{tab:data_assoc_stats}. G$k$CM correctly identifies
the 5 cliques corresponding to the different beacons and out performs
PCM in all the metrics.


\begin{table*}[tbh]
    \centering
    \vspace{3mm}
    \caption{Statistics for G$k$CM and PCM in Data Association experiment. Best results are in \textbf{BOLD}}

\begin{tabular}{|c|cc|cc|cc|cc|cc|ccc|}
\hline
\multirow{2}{*}{} & \multicolumn{2}{c|}{Translational RMSE (m)} & \multicolumn{2}{c|}{Rotational  RMSE (rad)} & \multicolumn{2}{c|}{Beacon Error (m)} & \multicolumn{2}{c|}{Residual}    & \multicolumn{2}{c|}{Inliers}   & \multicolumn{3}{c|}{Chi2}                                                         \\ \cline{2-14}
                  & \multicolumn{1}{c|}{Avg}      & Std         & \multicolumn{1}{c|}{Avg}      & Std         & \multicolumn{1}{c|}{Avg}    & Std     & \multicolumn{1}{c|}{Avg} & Std   & \multicolumn{1}{c|}{TPR} & FPR   & \multicolumn{1}{c|}{Avg} & \multicolumn{1}{c|}{Std} & \multicolumn{1}{l|}{Median} \\ \hline
G$k$CM            & \textbf{0.6259}                        & \textbf{0.5646}      & \textbf{0.2469}                        & \textbf{0.0629}      & \textbf{34.14}                       & 52.06   & \textbf{72.12}                    & \textbf{88.67} & 0.84                    & \textbf{0.008} & \textbf{1.05}                     & \textbf{1.26}                     & \textbf{0.306}                       \\ \hline
PCM               & 3.153                         & 3.611       & 0.4521                        & 0.2218      & 40.28                       & \textbf{51.03}   & 1010                     & 1037  & \textbf{0.95}                    & 0.017 & 13.87                    & 14.16                    & 29.28                       \\ \hline
\end{tabular}
\label{tab:data_assoc_stats}
\vspace{-4mm} 
\end{table*}

\subsection{Tuning Experiment}
\begin{figure}[!tbh]
    \vspace{-3mm}
    \centering
    \includegraphics[width=0.93\columnwidth, trim=0 0 0 25, clip]{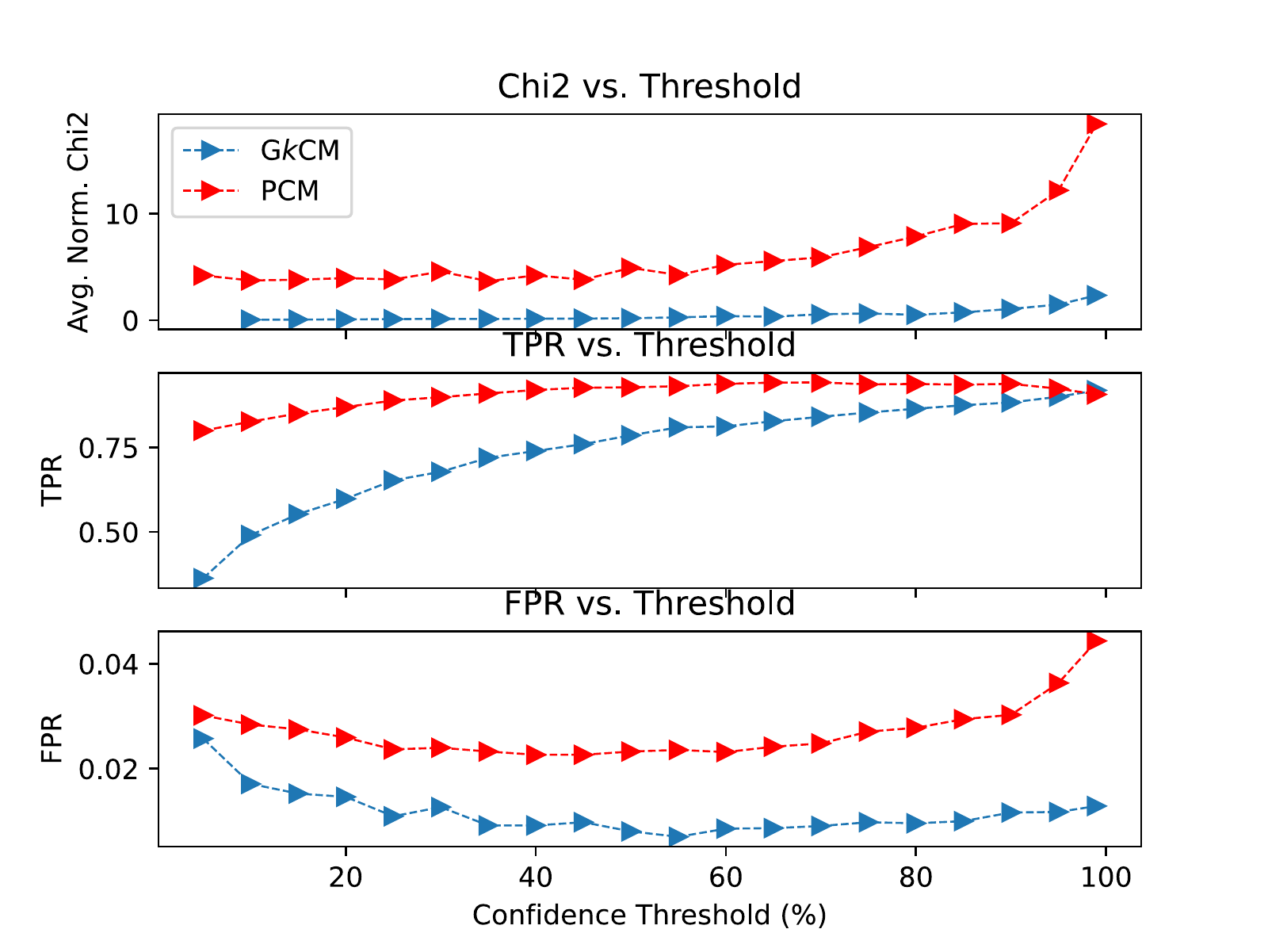}
    \caption{Results showing the normalized chi2 value, TPR and FPR by varying the consistency threshold value, $\gamma$, for a fixed trajectory.}
    \label{fig:varied_chi2}
\end{figure}

PCM has the nice property
that changing the threshold value, $\gamma$, did not
significantly impact the results of the algorithm. Due to enforced group
consistency as opposed to pairwise we designed an experiment to
test if G$k$CM has a similar property.
We accomplished this by fixing a robot trajectory of 50 poses and the
associated measurements and running G$k$CM multiple times with a different
value for $\gamma$ each time. The measurements contained 40 outliers
that were generated as described previously. We averaged the $\chi^2$ value and the true
and false positive rates over multiple runs. \Cref{fig:varied_chi2}
shows how the above values vary with the consistency threshold for both G$k$CM
and PCM.

As can be seen, G$k$CM performs better than PCM in both the normalized $\chi^2$ and
false positive rate, which is more important in our application than the true
positive rate. The results indicate that the performance of G$k$CM varies more
with the threshold $\gamma$ than results in \cite{mangelson2018pairwise},
especially at very low and high confidence thresholds. As such, we recommend
that confidence values be used from the $50-90\%$ confidence range where
performance was less variable with the confidence threshold.



\subsection{Incremental Update}
\begin{figure}[!tbh]
    \centering
    \includegraphics[width=0.93\columnwidth, trim=0 0 0 25, clip]{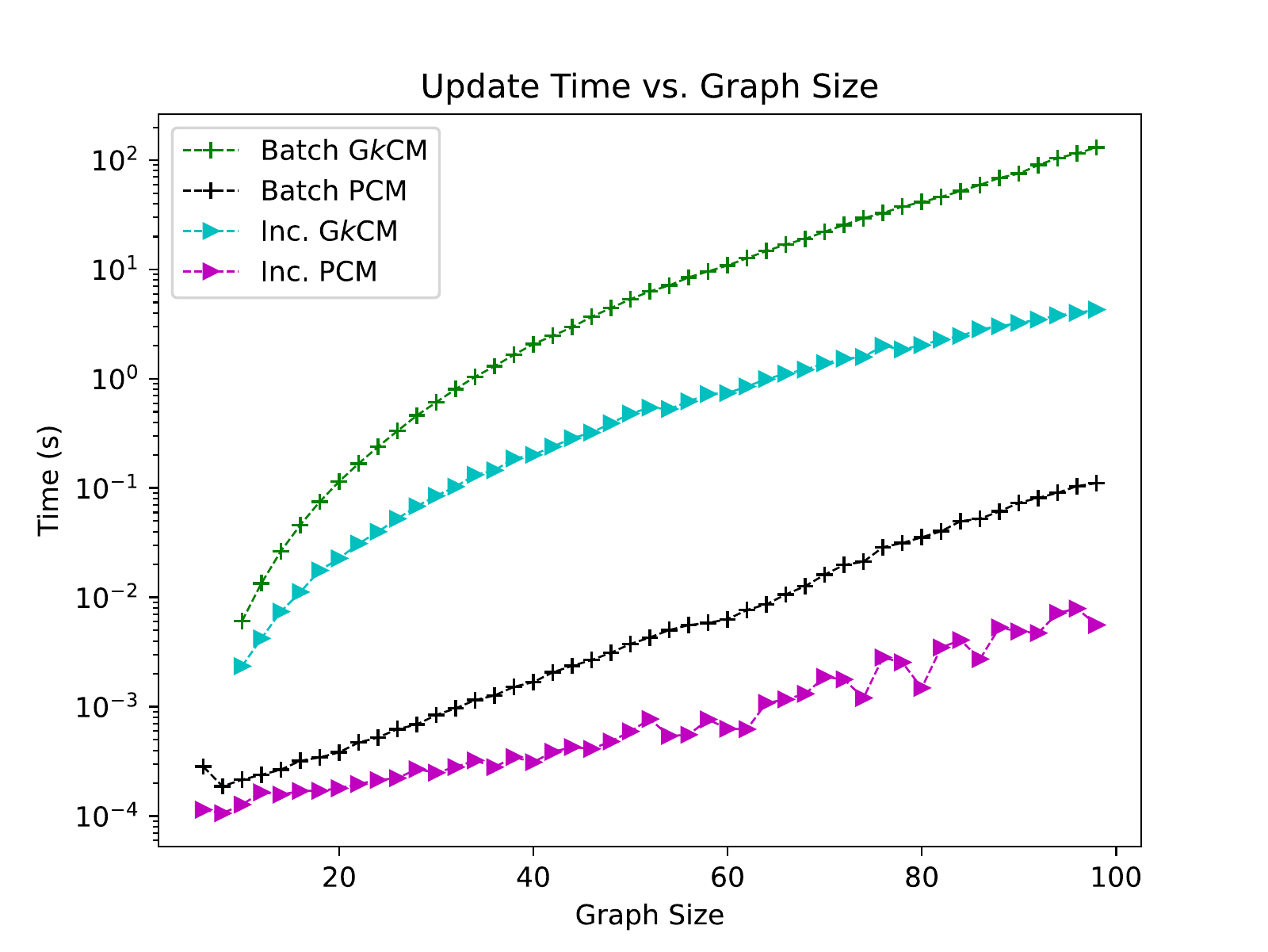}
    \caption{Timing data for both batch and incremental updates for G$k$CM and PCM. This includes the time to perform the relevant consistency checks and the new maximum clique. \emph{Note the log-scale on the vertical axis}.}
    \label{fig:timing_plot}
\end{figure}

In this last experiment we evaluate the incremental heuristic described in
\cite{chang2021kimera} since their experiments only evaluated the heuristic
for a $k$-uniform hypergraph where $k=2$. For this experiment we generate a
trajectory of 100 poses and measurements and at each step we evaluate how
long both an incremental and batch update take. Updates include
performing the consistency checks and finding the maximum clique.
We record the runtime for the graph size and average statistics over multiple
runs. We plot the runtime against the size of the graph in
\cref{fig:timing_plot}.

As can be seen the incremental update with the heuristic in \cite{chang2021kimera}
provides similar benefits for G$k$CM as it does for PCM.
On average, for a graph of 100 nodes with 80 outliers, it takes a batch
solution over 40 seconds to solve for the maximum clique while it takes only
3 seconds for the incremental update. These findings
validate the results in \cite{chang2021kimera} and also allow for G$k$CM to be
run closer to real time.

\section{Conclusion}
\label{section:conclusion}

In this paper we introduced a novel concept called group-$k$ consistency
maximization or G$k$CM. By modifying existing maximum clique algorithms to work
over generalized graphs we can select groups of consistent measurements in
high outlier regimes where pairwise consistency is inadequate.




\bibliographystyle{style/IEEEtran}
\bibliography{IEEEabrv, strings-short, library}

\end{document}